\documentclass[letterpaper, 10 pt, conference]{ieeeconf}  

\IEEEoverridecommandlockouts                              

\usepackage{epsfig} 
\usepackage{mathptmx} 
\usepackage{times} 
\usepackage{amsmath} 
\usepackage{amssymb}  
\usepackage{graphicx}
\usepackage{hyperref}
\usepackage{xcolor}       
\usepackage{colortbl}     
\usepackage[RGB]{xcolor}  
\definecolor{mypink}{RGB}{255, 139, 139}
\definecolor{mycyan}{RGB}{0, 189, 204}
\definecolor{myblue}{RGB}{91, 155, 213}
\definecolor{mygreen}{RGB}{169, 209, 142}
\definecolor{mybluebg}{RGB}{197,234,251}
\definecolor{mygraybg}{RGB}{235,235,235}

\title{\LARGE \bf
Fusion4CA: Boosting 3D Object Detection via Comprehensive Image Exploitation
}

\author{
Kang Luo$^*$, Xin Chen$^*$, Yangyi Xiao, Hesheng Wang$^\dag$
\thanks{Kang Luo, Xin Chen, Yangyi Xiao and Hesheng Wang are with IRMV Lab, the Department of Automation, Shanghai
Jiao Tong University.}
\thanks{*Equal contribution}
\thanks{$^\dag$Corresponding author email: wanghesheng@sjtu.edu.cn}
}

\begin{document}
\maketitle
\thispagestyle{empty}
\pagestyle{empty}

\begin{abstract}

Nowadays, an increasing number of works fuse LiDAR and RGB data in the bird's-eye view (BEV) space for 3D object detection in autonomous driving systems. However, existing methods suffer from over-reliance on the LiDAR branch, with insufficient exploration of RGB information. To tackle this issue, we propose Fusion4CA, which is built upon the classic BEVFusion framework and dedicated to fully exploiting visual input with plug-and-play components. Specifically, a contrastive alignment module is designed to calibrate image features with 3D geometry, and a camera auxiliary branch is introduced to mine RGB information sufficiently during training. For further performance enhancement, we leverage an off-the-shelf cognitive adapter to make the most of pre-trained image weights, and integrate a standard coordinate attention module into the fusion stage as a supplementary boost. Experiments on the nuScenes dataset demonstrate that our method achieves 69.7\% mAP with only 6 training epochs and a mere 3.48\% increase in inference parameters, yielding a 1.2\% improvement over the baseline which is fully trained for 20 epochs. Extensive experiments in a simulated lunar environment further validate the effectiveness and generalization of our method. Our code will be released through \href{https://github.com/Gorgeousful/Fusion4CA}{Fusion4CA}.

\end{abstract}

\section{INTRODUCTION}

3D Object detection is an indispensable module in modern autonomous driving systems, which demands reliable recognition, precise 3D localization, and accurate geometry estimation of complex targets in dynamic driving scenarios \cite{yurtsever2020survey,arnold2019survey}. LiDAR has been the primary sensor for mainstream 3D detection pipelines \cite{wu2022sparse,zhang2024safdnet,chen2023voxelnext}, but its performance is inevitably constrained by inherent bottlenecks, including the sparsity of raw point clouds, sensitivity to the reflectivity of the surface, and performance degradation in adverse weather \cite{wu2023virtual,li2020lidar}. To mitigate these limitations, a mainstream research paradigm focuses on fusing RGB data captured by on-board cameras, leveraging their dense texture and rich semantic information to complement LiDAR measurements and further enhance detection performance \cite{ibrahim2025multistream,bai2022transfusion,guan2022m3detr}.

If taking one modality as the dominant one and embedding the features of the other modality into it, the final fused representation will be inherently constrained by the intrinsic characteristics of the primary modality \cite{liu2023bevfusion}. Consequently, how to effectively fuse the texture and semantic advantages of images with the spatial geometric advantages of LiDAR has become a key research priority. Recently, the BEV-based perception method has become the mainstream fusion paradigm for Camera-LiDAR-based 3D object detection \cite{liu2023bevfusion, liang2022bevfusion}, due to its unified view representation and natural compatibility with downstream tasks in autonomous driving.

However, most existing BEV-based approaches still suffer from an excessive reliance on the LiDAR modality, with insufficient exploitation of the camera modality \cite{liu2023bevfusion, zhao2024simplebev}. This critical drawback results in only marginal performance improvements of multi-modal fusion schemes compared with LiDAR-only detection methods. We attribute this long-standing performance bottleneck to the following points: (1) The encoded image features are not geometrically calibrated before entering the view transform stage; (2) The standalone supervision signal struggles to effectively guide the optimization of the camera branch when LiDAR information alone is sufficient to accomplish most tasks; (3) Full-parameter fine-tuning fails to fully unleash the representation potential of pre-trained weights from the image encoder due to large-scale networks; (4) The fusion module lacks an efficient mechanism to capture discriminative information from each individual modality.

\begin{figure*}[tb] 
    \centering
    \includegraphics[width=\linewidth]{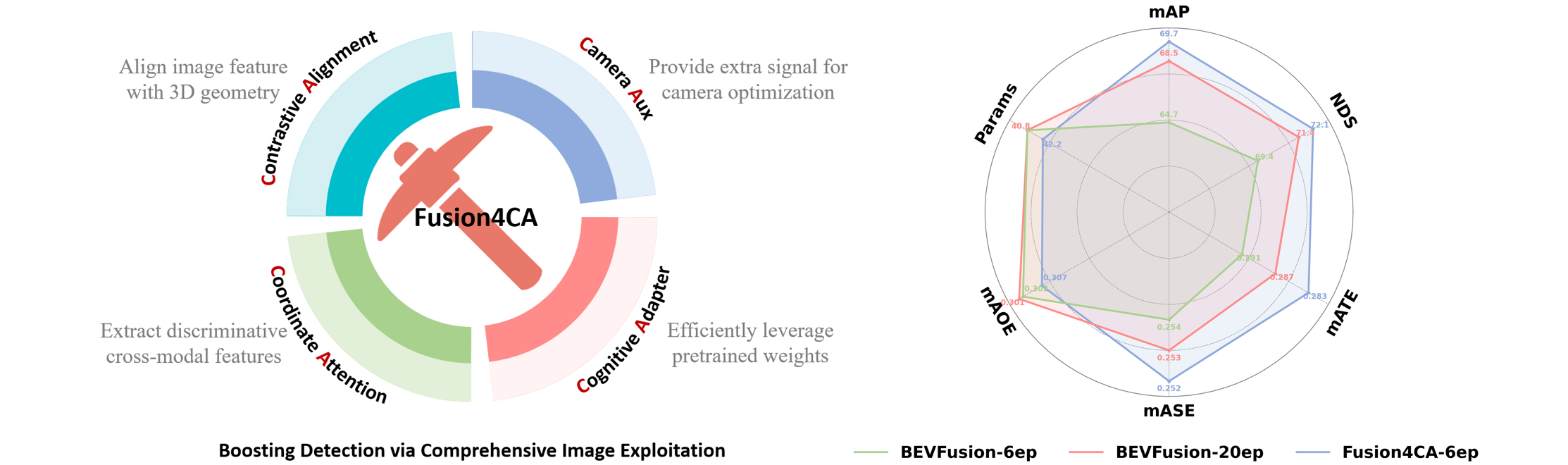}
    \caption{
    Key components of our Fusion4CA framework, consisting of Contrastive Alignment Module, Camera Auxiliary Branch, Cognitive Adapter and Coordinate Attention Module.
    Our model outperforms BEVFusion by 5\% mAP at six epochs and surpasses its 20-epoch counterpart by 1.2\% mAP.}
    \label{fig:Design}
\end{figure*}

 In this work, we propose Fusion4CA, an improved camera-LiDAR fusion framework built upon BEVFusion \cite{liu2023bevfusion} to better exploit visual information. As illustrated in Fig.~\ref{fig:Design}, we introduce four complementary components to alleviate the over-reliance on the LiDAR modality and fully unlock the potential of RGB data. Specifically, a Contrastive Alignment Module is designed to perform calibration on the encoded image features before they enter the view transform stage, ensuring the alignment between image features and 3D spatial structure. To tackle the insufficient guidance of standalone supervision signals under LiDAR dominance, we propose a Camera Auxiliary Branch, which provides additional supervision for the optimization of the camera branch, promoting the full exploration of texture and semantic information. We further adopt an off-the-shelf Cognitive Adapter \cite{yin20255} to effectively utilize pre-trained image weights, and integrate a standard Coordinate Attention Module \cite{hou2021coordinate} to capture discriminative cross-modal features. Notably, all these components are \textit{plug-and-play} and can be readily integrated into other baseline frameworks. Our contributions are as follows:

\begin{itemize}
\item We propose Fusion4CA, an effective Camera-LiDAR fusion framework built upon BEVFusion, which alleviates the over-dependence on LiDAR signals and fully exploits the representation power of RGB images for 3D Object Detection.
\item We design a Contrastive Alignment Module to enforce alignment between visual features and 3D spatial geometry, together with a Camera Auxiliary Branch that provides extra supervision to mitigate the LiDAR-dominated training bias and enhance the exploitation of image texture and semantics.
\item Our method achieves competitive 3D detection performance on the nuScenes dataset with only 6 training epochs and negligible extra inference overhead, while promising results on our custom-built simulated lunar environment further validate its effectiveness and strong generalization capability.
\end{itemize}

\section{RELATED WORK}

\subsection{3D Object Detection with Camera Modality}
Mainstream approaches for camera-based 3D object detection can generally be divided into depth-based methods and network-based methods. Depth-based schemes \cite{li2023bevdepth,philion2020lift,lu2025toward} explicitly estimate depth and project image features into BEV space with camera parameters. Nevertheless, such methods are highly dependent on implicit depth estimation, which tends to suffer performance degradation in ambiguous depth scenarios, especially for distant objects and texture-less regions.
By contrast, network-based methods \cite{li2024bevformer,yang2023bevformer,jiang2023polarformer} implicitly lift image features to the BEV space through neural networks, typically Transformers. Despite recent progress, these approaches still exhibit obvious limitations. They require large-scale training data and massive computational resources for stable convergence, and full-parameter fine-tuning of Transformer structures also introduces excessive GPU memory overhead and high training costs \cite{yin20255}.

\subsection{3D Object Detection with LiDAR Modality} 
LiDAR-based 3D object detection methods are mainly categorized into point-based approaches and grid-based approaches according to the point cloud feature extraction paradigm. Point-based methods \cite{qi2017pointnet,qi2017pointnet++,ding2019votenet} operate directly on raw LiDAR point clouds by exploiting the unordered nature of point sets to capture geometric information with max pooling. Alternatively, grid-based methods \cite{zhou2018voxelnet,lang2019pointpillars,yan2018second} first partition the LiDAR point cloud into pre-defined regular voxels or pillars and then apply convolutions on the grid representation. However, such methods are limited by the inherent properties of point clouds, whose features are often sparse and sensitive to object surface reflectance and adverse weather conditions.

\subsection{3D Object Detection with Multi-Modalities} 
3D perception via multi-modal fusion can be categorized into three paradigms based on the type of fused features: primary-auxiliary modality fusion (with either image or point cloud as the primary modality), BEV-based feature fusion, and Query-based fusion.
The primary-auxiliary paradigm enhances the primary modality with complementary information from the auxiliary modality, and performs final 3D detection on the primary features. However, its final performance is constrained by the inherent limitations of the primary modality, such as the sparsity of the point cloud \cite{ibrahim2025multistream} or insufficient geometric information \cite{hu2023ea, qi2018frustum,vora2020pointpainting}.
The BEV-based approach \cite{liu2023bevfusion,liang2022bevfusion,cai2023bevfusion4d} projects camera images and LiDAR point clouds into the BEV space for subsequent processing. However, projecting image features into BEV space tends to cause information loss, and it is difficult to effectively supervise the camera branch under large-scale network settings and LiDAR-dominated training.
The Query-based approach \cite{bai2022transfusion,zhang2024sparselif,wang2025mv2dfusion} comprehensively fuses LiDAR and image information via the Transformer attention mechanism. However, it relies heavily on large-scale training data and is prone to overfitting under sparse data or domain shift scenarios.

\begin{figure*}[tb] 
    \centering
    \includegraphics[width=0.9\textwidth]{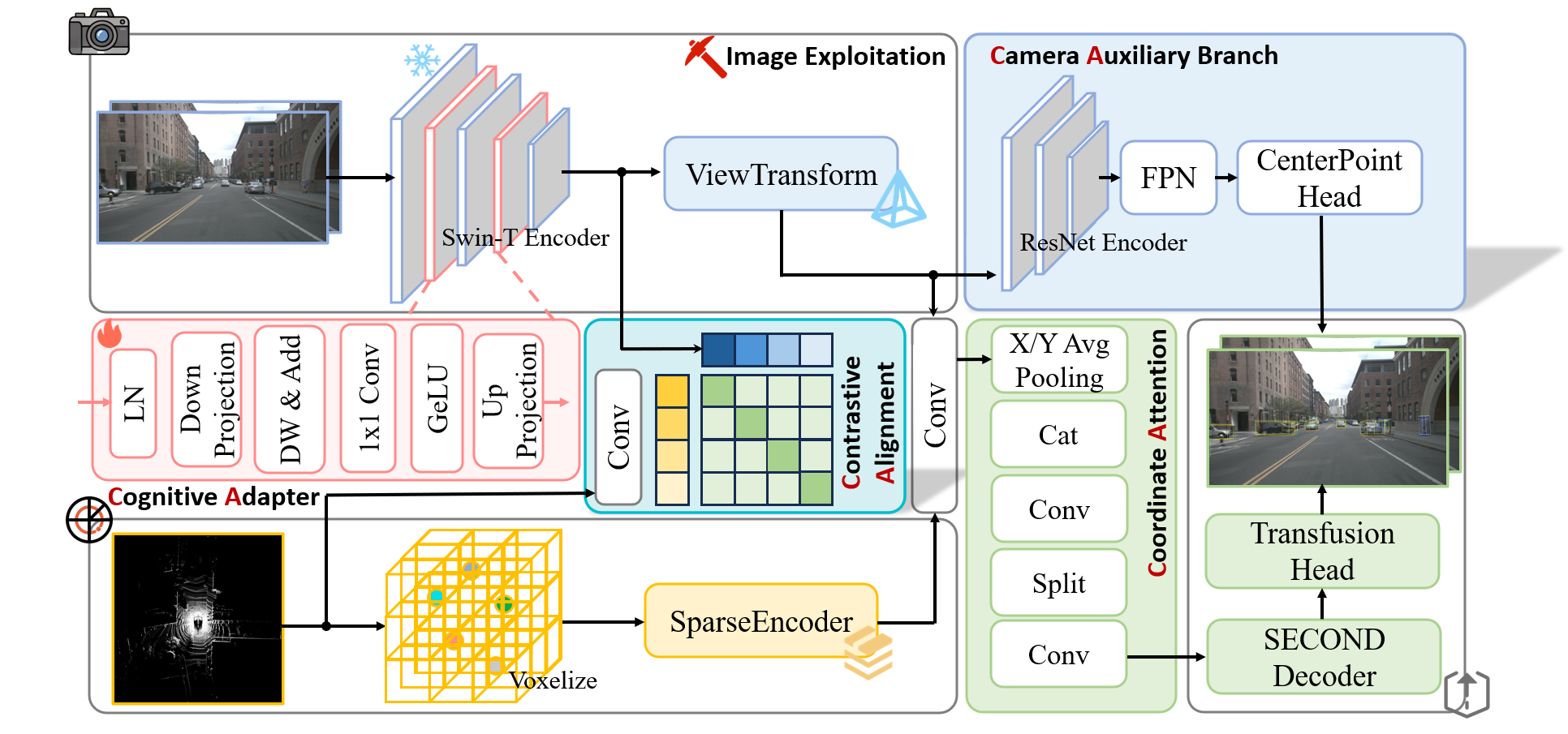}
    \caption{
    An overview of the proposed Fusion4CA network with four plug-and-play enhancements for visual exploitation.
    (1) A Contrastive Alignment Module is designed to align image features with projected point cloud features.
    (2) A Camera Auxiliary Branch is proposed to provide extra supervision for direct optimization of the camera branch.
    (3) An off-the-shelf Cognitive Adapter is inserted into the Swin Transformer while keeping its original weights frozen.
    (4) A standard Coordinate Attention Module is appended after convolutional fusion to capture discriminative information effectively.
    Note that residual connections are omitted for brevity.
    }
    \label{fig:Pipeline}
\end{figure*}

\section{METHODOLOGY}
The overall pipeline of the proposed method is illustrated in Fig.~\ref{fig:Pipeline}. Built upon BEVFusion \cite{liu2023bevfusion}, our framework integrates four \textit{plug-and-play} components to fully exploit the potential of RGB images and enhance cross-modal feature fusion.
The network first extracts multi-modal features using respective backbones.
The image features are then converted into image-BEV representations, where the Contrastive Alignment Module is employed to achieve explicit feature alignment.
The image-BEV features are subsequently fused with the LiDAR-BEV features, and the Coordinate Attention Module \cite{hou2021coordinate} is adopted to capture discriminative multi-modal representations.
The refined features are then fed into the decoder and detection head to produce final results.
Specifically, we insert the Cognitive Adapter \cite{yin20255} into the camera backbone, freeze the pre-trained weights during backpropagation, and update only a small number of parameters in the adapter, enabling efficient tuning with enhanced performance.
Furthermore, the Contrastive Alignment Module and Camera Auxiliary Branch are activated only during training, enabling the network to perform inference with negligible additional parameters.
We will elaborate on the details of each key component in the following sections.

\begin{figure}[tb] 
    \centering
        \includegraphics[width=0.9\linewidth]{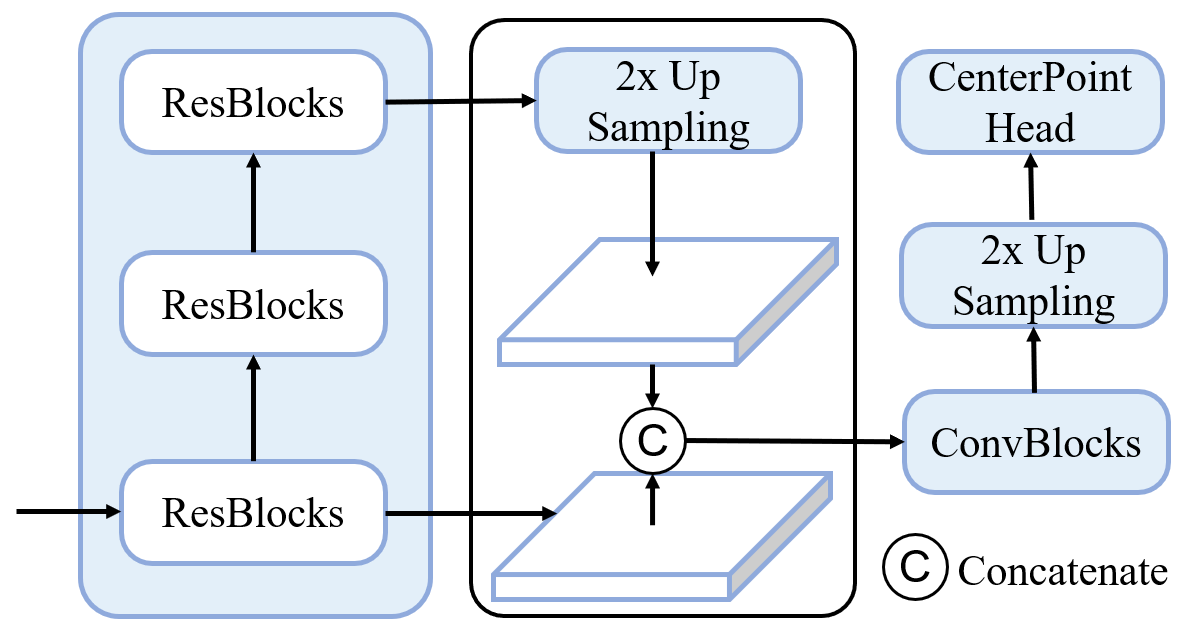}
        \caption{Illustration of the Camera Auxiliary Branch, comprising stacked residual blocks, FPN, and CenterPoint Head. 
        The primary function is to provide supplementary supervision signals to directly optimize the camera branch.}
        \label{fig:Camera_Aux}
\end{figure}

\subsection{Contrastive Alignment for Image Calibration}
In the baseline model, the LiDAR branch and the camera branch are relatively independent and have no interaction before convolutional fusion, leading to insufficient multi-modal interaction. Meanwhile, features from the image encoder lack effective geometric alignment before entering the view transform, which directly affects the subsequent forward propagation. In order to solve that, we introduce a Contrastive Alignment Module before the view transform to provide extra supervision signals during training, align RGB features with point cloud features, and preserve their semantic consistency. The module is simple yet effective as illustrated in Fig.~\ref{fig:Pipeline}.

We employ temperature-scaled cross-entropy loss \cite{chen2020simple} as the core of the Contrastive Alignment Module. 
This loss maximizes the similarity between RGB–depth feature pairs from the same sample and camera view, and enlarges the discrepancy between those from different samples or distinct camera views.
First, we preprocess the RGB and depth features to ensure that their flattened vectors share the same length. Specifically, a three-layer convolutional block is used to gradually align the channel number of depth features with image features.
\begin{equation}
        \displaystyle
        \left\{
        \begin{aligned}
            & x_{rgb} = Flat(x_{rgb}^0)\\
            & x_{dep} = Flat(Conv(x_{dep}^0))
        \end{aligned}
        \right.
\end{equation}

We then compute the cross-entropy loss based on $x_{rgb}$ and $x_{dep}$, which can be formulated as follows, where the hyperparameter $\tau$ controls the sharpness of alignment and $B$ represents the batch size.
\begin{equation}
        \displaystyle
        \left\{
        \begin{aligned}
            & L_{align}=-\dfrac{1}{B}\sum_{i=1}^{B}\log\dfrac{\exp({sim(x_{rgb}^i, x_{dep}^i)/\tau})}{\sum_{j=1}^{B}\exp({sim(x_{rgb}^i, x_{dep}^j)/\tau})}\\
            & sim(x_{rgb}^i,x_{dep}^j)=\dfrac{x_{rgb}^i \cdot x_{dep}^j}{||x_{rgb}^i||\;||x_{dep}^j||}
        \end{aligned}
        \right.
\end{equation}

\subsection{Camera Auxiliary Branch for Visual Supervision}
In order to tackle the insufficient guidance of standalone supervision signals under LiDAR dominance, we design a Camera Auxiliary Branch to provide additional supervision signals to directly optimize the camera side.  Figure~\ref{fig:Camera_Aux} illustrates the structure of the auxiliary branch. The structure of the branch is relatively simple: we first use three stacked residual blocks to compress the features from the camera branch. Then, an FPN-like structure is adopted to perform feature fusion. Finally, supervision is achieved through a CenterPoint detection head \cite{yin2021center} with auxiliary loss $L_{aux}$, whose calculation process is consistent with that of the main branch \cite{liu2023bevfusion} and calculated merely in the training phase.

\subsection{Image Encoder Enhanced by Cognitive Adapter} 

\begin{figure}[tb] 
    \centering
        \includegraphics[width=0.8\linewidth]{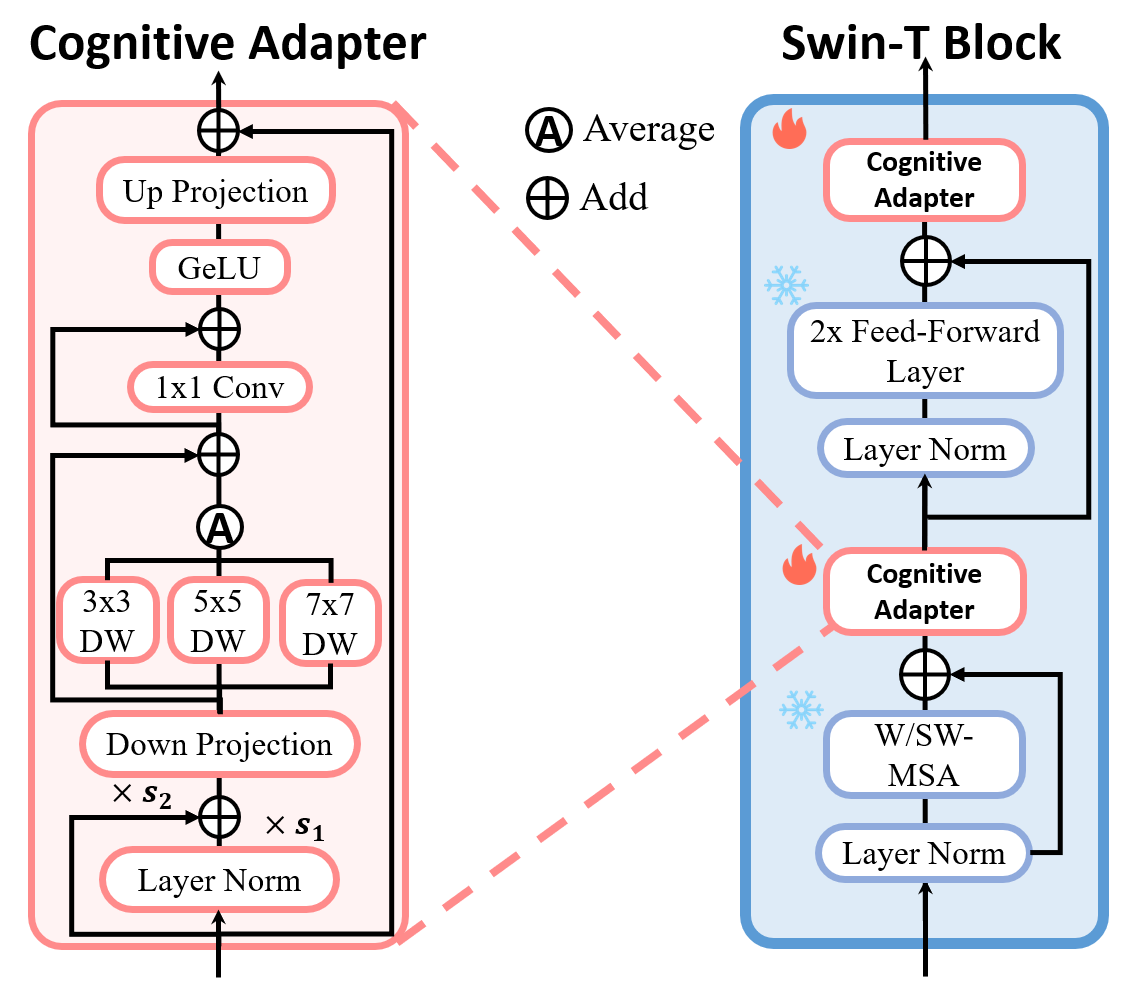}
        \caption{The Cognitive Adapter is inserted after the self-attention and feed-forward layers in each Swin-T block, where adaptive layer normalization, depthwise convolution and residual connections are employed to boost feature expressiveness.}
        \label{fig:Cognitive_Adapter}
\end{figure}

As depicted in Fig.~\ref{fig:Cognitive_Adapter}, the Cognitive Adapter \cite{yin20255} is integrated into each Swin-Transformer block. In order to unleash the representation potential of the image encoder, the model is optimized via delta tuning. In contrast to full fine-tuning, delta tuning only requires fine-tuning a small number of parameters in the added lightweight module, drastically cutting down training costs while preserving the general knowledge encoded in the pre-trained weights.
Given the input feature $x_{img}^l $ of the Swin-T backbone in stage $l$ , the processing procedure within adapter can be formulated as follows:
\begin{equation}
        \displaystyle
        \left\{
        \begin{aligned}
            & x_{\text{img}}^{l+1} = x_{\text{img}}^l + U_l \sigma\!\left(f_{\text{pw}}\!\left(f_{\text{dw}}\!\left(D_l\!\left(x_{\text{norm}}^l\right)\right)\right)\right) \\
            & x_{\text{norm}}^l = s_1 \cdot LN\!\left(x_{\text{img}}^l\right) + s_2 \cdot x_{\text{img}}^l
        \end{aligned}
        \right.
\end{equation}

Here, $\sigma(\cdot)$ denotes GeLU activation. $LN(\cdot)$ represents Layer Normalization, while $U(\cdot)$ and $D(\cdot)$ represent the upward projection and downward projection. Additionally, $f_{dw}$ denotes multi-scale depthwise convolution (with residual connections) and  $f_{pw}$ stands for $1\times1$ convolution, while $s_1$ and $s_2$ are trainable scaling factors. 

\subsection{Fusion Refinement with Coordinate Attention}
\begin{figure}[tb] 
    \centering
        \includegraphics[width=\linewidth]{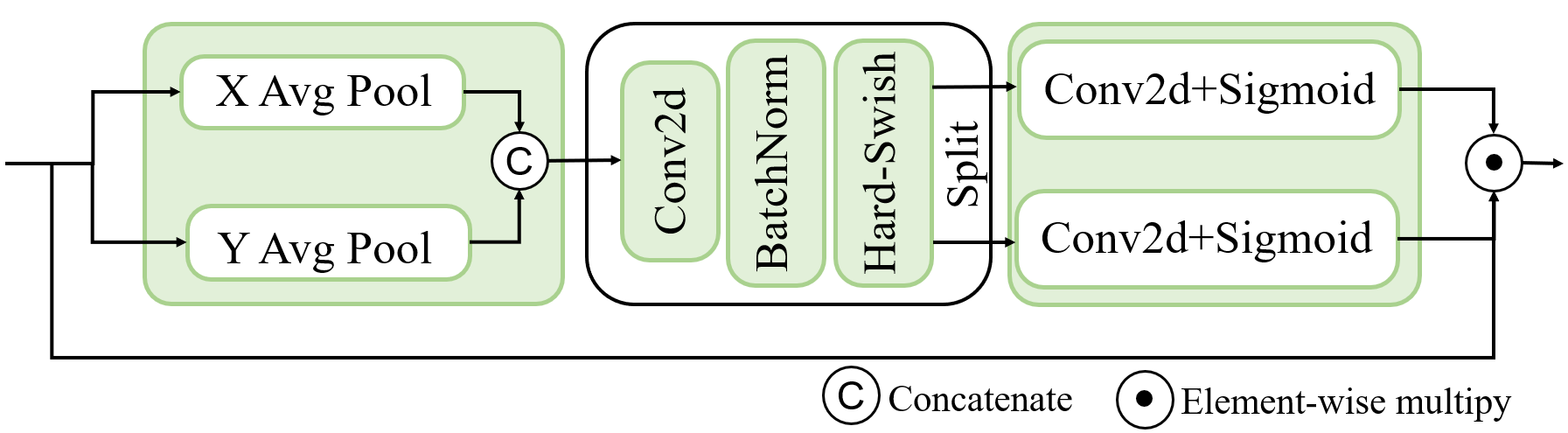}
        \caption{Illustration of Coordinate Attention Module. 
        The module applies 1D global average pooling along two directions to compute direction-sensitive attention weights, then enhances the input via element-wise multiplication and a residual connection.
        }
        \label{fig:Coordinate_Attn}
\end{figure}

We append a Coordinate Attention Module \cite{hou2021coordinate} behind the convolutional fusion to capture discriminative information from multi-modal features. The structure of the coordinate attention module is illustrated in Fig.~\ref{fig:Coordinate_Attn}. The module first performs 1D global average pooling on the input along the horizontal and vertical directions, respectively, to generate direction-aware intermediate features. It then concatenates the features from the two directions and applies a non-linear transformation after a shared $1\times1$ convolution. Subsequently, it splits the fused features into horizontal and vertical components, which are individually activated by the sigmoid function to generate direction-sensitive channel attention weights. Finally, through residual connection, the attention maps from the two directions are multiplied element-wise by the original input to produce the features enhanced by coordinate attention.

\begin{table*}[htbp]
    \centering
    \caption{Comparison on the nuScenes dataset. 
    ‘C.V.’, ‘T.L.’, ‘B.R.’, ‘M.T’, ‘Ped.’ and ‘T.C.’ are short for construction vehicle, trailer, barrier, motor, pedestrian and traffic cone, respectively. ‘L’ and ‘C‘ are short for LiDAR and camera. Note that our method only trained for 6 epochs, while others are fully trained.}
    \label{tab:Comparision}
    \begin{tabular}{c|c|c|cc|cccccccccc}
    \noalign{\hrule height 1pt}
    Method &  Reference & Mod. & mAP & NDS & Car & Truck & C.V. & Bus & T.L. & B.R. & M.T. & B.C. & Ped. & T.C. \\
    \hline
    \multicolumn{14}{c}{Results on the validation data set} \\
    \hline
    BEVFusion \cite{liu2023bevfusion} & ICRA 2023 & L & 64.7 & 69.3 & 86.9 & 61.0 & 27.3 & 72.5 & 41.8 & 69.6 & 71.7 & 56.3 & 86.6 & 73.2 \\
    BEVFusion \cite{liu2023bevfusion} & ICRA 2023 & L+C & 68.5 & 71.4 & 89.2 & 64.6 & 30.4 & 75.4 & 42.5 & 72.0 & 78.5 & 65.3 & 88.2 & 79.5 \\
    \rowcolor{mybluebg}
    \textbf{Fusion4CA (Ours)} & - & L+C & \textbf{69.7} &\textbf{ 72.1} & \textbf{89.7} & \textbf{66.2} & \textbf{31.9} & \textbf{77.3} & \textbf{43.6} & \textbf{72.3} & \textbf{79.5} & \textbf{66.3} & \textbf{89.5} & \textbf{80.3} \\
    \hline
    \multicolumn{14}{c}{Results on the test data set} \\
    \hline
    CenterPoint \cite{yin2021center} & CVPR 2021 & L & 60.3 & 67.3 & 85.2 & 53.5 & 20.0 & 63.6 & 56.0 & 71.1 & 59.5 & 30.7 & 84.6 & 78.4 \\
    Focals Conv \cite{chen2022focal} & CVPR 2022 & L & 63.8 & 70.0 & 86.7 & 56.3 & 23.8 & 67.7 & 59.5 & 74.1 & 64.5 & 36.3 & 87.5 & 81.4 \\
    TransFusion-L \cite{bai2022transfusion} & CVPR 2022 & L & 65.5 & 70.2 & 86.2 & 56.7 & 28.2 & 66.3 & 58.8 & \textbf{78.2} & 68.3 & 44.2 & 86.1 & 82.0 \\
    VoxelNeXt \cite{chen2023voxelnext} & CVPR 2023 & L & 66.2 & 71.4 & 85.3 & 55.7 & 29.8 & 66.2 & 57.2 & 76.1 & 75.2 & 48.8 & 86.5 & 80.7 \\
    \hline
    MVP \cite{yin2021multimodal} & NeurIPS 2021 & L+C & 66.4 & 70.5 & 86.8 & 58.5 & 26.1 & 67.4 & 57.3 & 74.8 & 70.0 & 49.3 & 89.1 & 85.0 \\
    GraphAlign \cite{song2023graphalign} & ICCV 2023 & L+C & 66.5 & 70.6 & 87.6 & 57.7 & 26.1 & 66.2 & 57.8 & 74.1 & 72.5 & 49.0 & 87.2 & 86.3 \\
    PointAugmenting \cite{wang2021pointaugmenting} & CVPR 2021 & L+C & 66.8 & 71.0 & 87.5 & 57.3 & 28.0 & 65.2 & 60.7 & 72.6 & 74.3 & 50.9 & 87.9 & 83.6 \\
    FusionPainting \cite{xu2021fusionpainting} & ITSC 2021 & L+C & 68.1 & 71.6 & 87.1 & 60.8 & 30.0 & 68.5 & 61.7 & 71.8 & \textbf{74.7} & 53.5 & 88.3 & 85.0 \\
    TransFusion \cite{bai2022transfusion} & CVPR 2022 & L+C & 68.9 & 71.7 & 87.1 & 60.0 & 33.1 & 68.3 & 60.8 & 78.1 & 73.6 & 52.9 & 88.4 & \textbf{86.7} \\
    BEVFusion \cite{liang2022bevfusion} & NeurIPS 2022 & L+C& 69.2 & 71.8 & 88.1 & 60.9 & 34.4 & 69.3 & 62.1 & \textbf{78.2} & 72.2 & 52.2 & 89.2 & 85.2 \\
    FUTR3D \cite{chen2023futr3d} & CVPR 2023 & L+C & 69.4 & \textbf{72.1} & 86.3 & \textbf{61.5} & 26.0 & 71.9 & 42.1 & 64.4 & 73.6 & \textbf{63.3} & 82.6 & 70.1 \\
    \rowcolor{mybluebg}
    \textbf{Fusion4CA (Ours)} & - & L+C & \textbf{69.7} & \textbf{72.1} & \textbf{88.7} & 61.4 & \textbf{36.6} & \textbf{72.4} & \textbf{63.5} & 74.5 & 74.3 & 50.1 & \textbf{89.3} & 86.4 \\
    \noalign{\hrule height 1pt}
    \end{tabular}
\end{table*}


\begin{table*}[htbp]
    \centering
    \caption{Ablation study on nuScenes validation set using different component combinations.}
    \label{tab:Ablation}
    \begin{tabular}{c|cccc|cc|cc|ccc}
        \noalign{\hrule height 1pt}
        Order & \textcolor{mycyan}{ConAlign} & \textcolor{myblue}{CamAux} & \textcolor{mygreen}{CoordAtt} & \textcolor{mypink}{CogAdp} & mAP & $\Delta$mAP & NDS & $\Delta$NDS & mATE & mASE & mAOE \\
        \hline
        01 & & & & & 64.7 & - & 69.4 & - & 0.291 & 0.254 & 0.302\\
        02 & $\checkmark$ & & & & 67.0 & +2.3 & 70.4 & +1.0 & 0.291 & 0.256 & 0.330 \\
        03 & & $\checkmark$ & & & 68.7 & +4.0 & 71.5 & +2.1 & 0.285 & 0.256 & 0.308 \\
        04 & & & $\checkmark$ & & 64.6 & -0.1 & 69.4 & +0.0 & 0.297 & 0.255 & \textbf{0.294} \\
        \hline
        05 & $\checkmark$ & $\checkmark$ & & & 68.9 & +4.2 & 71.5 & +2.1 & \textbf{0.281} & 0.255 & 0.319 \\
        06 & $\checkmark$ & $\checkmark$ & $\checkmark$ & & 69.3 & +4.6 & 71.7 & +2.3 & 0.287 & 0.256 & 0.315 \\
        \rowcolor{mybluebg}
        07 & $\checkmark$ & $\checkmark$ & $\checkmark$ & $\checkmark$ & \textbf{69.7} & \textbf{+5.0} & \textbf{72.1} & \textbf{+2.7} & 0.283 & \textbf{0.252} & 0.307 \\
        \noalign{\hrule height 1pt}
    \end{tabular}
\end{table*}

\begin{figure}[htb] 
    \centering
        \includegraphics[width=\linewidth]{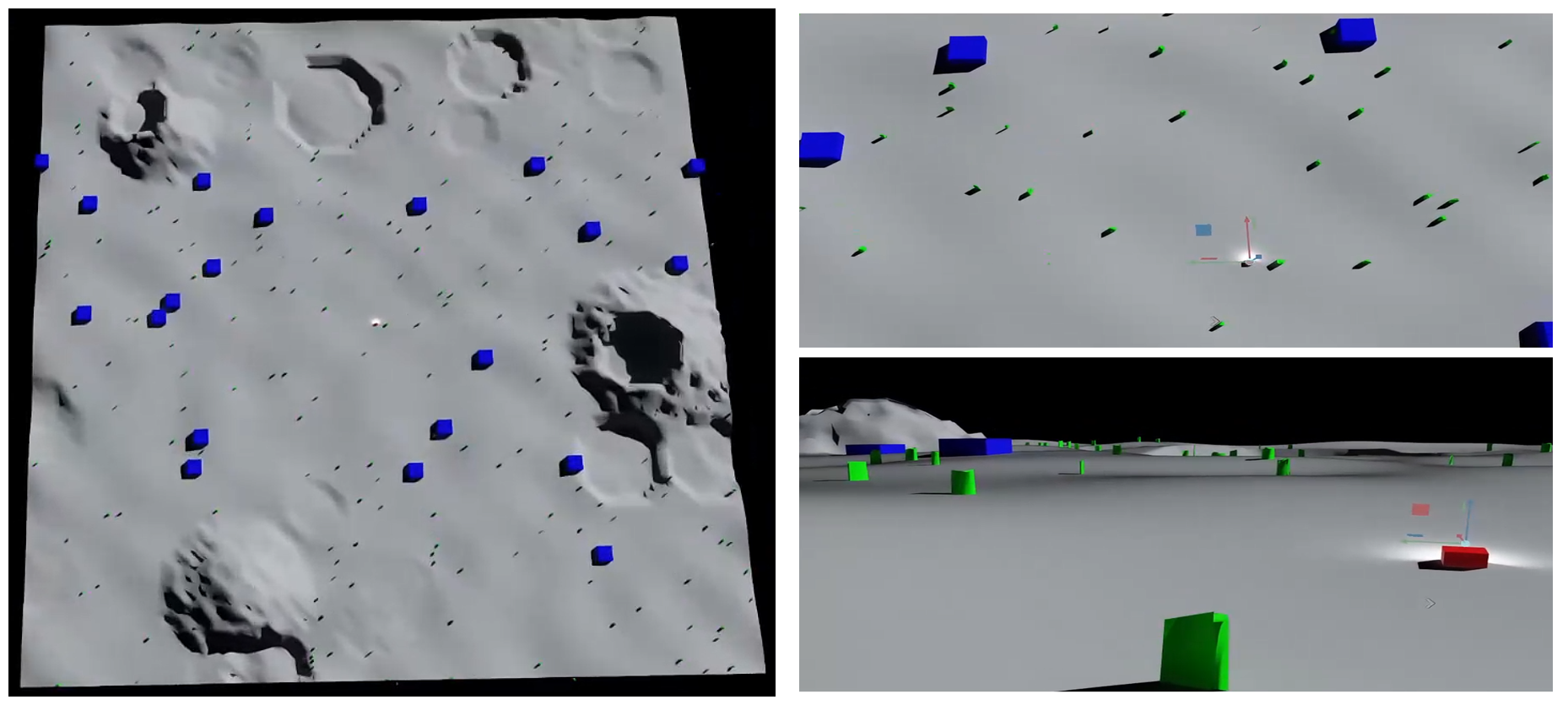}
        \caption{The simulated lunar environment in NVIDIA Isaac Sim, which is characterized by uneven terrain and craters with multiple protrusions and depressions. There are two categories to detect: Meteor (green) and Platform (blue). 
        The gray appearance of Meteors (green here only for visualization) is similar to lunar surface, posing significant challenges for the camera branch.}
        \label{fig:Sim_Env}
\end{figure}

\section{EXPERIMENTS}

\subsection{Experimental Setup}
\textbf{Datasets.} Experiments were conducted on the nuScenes \cite{caesar2020nuscenes} dataset and a photorealistic lunar-like simulation environment built in NVIDIA Isaac Sim. 
The nuScenes dataset provides 32-beam LiDAR point clouds (20 Hz) and RGB images from 6 surrounding cameras (12 Hz, 1600$\times$900 resolution), comprising 1000 annotated scenes covering 10 object categories. These scenes are split into training/validation/test subsets with a ratio of 700/150/150.

Additionally, as shown in Fig.~\ref{fig:Sim_Env}, the simulated lunar environment is characterized by uneven terrain and craters with multiple protrusions and depressions, and includes two object categories: Meteor (small, irregular-shaped) and Platform (large, regular-shaped). And the inspection robot deployed in this environment is equipped with a 32-channel LiDAR (10 Hz), an RGB camera (1900$\times$1200 resolution, 10 Hz) and an odometer (20 Hz).
Considering lunar illumination conditions, we configured two lighting setups and collected 5 ROS bag files for each setup, with each file lasting 5 minutes and a total data volume of 200 GB.
We randomly selected one ROS bag from each lighting group as test set and used the remaining bags for training.

\textbf{Implementation Details.} Our method is implemented based on the BEVFusion codebase \cite{liu2023bevfusion}. 
The model is trained for only 6 epochs with a batch size of 6 and an initial learning rate of 2e-4, using two RTX 4090 GPUs.
The Contrastive Alignment Module and Camera Auxiliary Branch are employed only for training and omitted during inference. Besides, the remaining modules introduce merely a total of 3.48\% increase in inference parameters.
Without test-time augmentation (TTA) or model ensemble, evaluations on the nuScenes validation set and the simulated lunar environment test set are conducted locally, while metrics on the nuScenes test set are evaluated via the EvalAI server \cite{yadav2019evalai}.

\subsection{Multi-class Results on nuScenes Dataset}
\begin{figure*}[htb] 
    \centering
        \includegraphics[width=0.95\linewidth]  {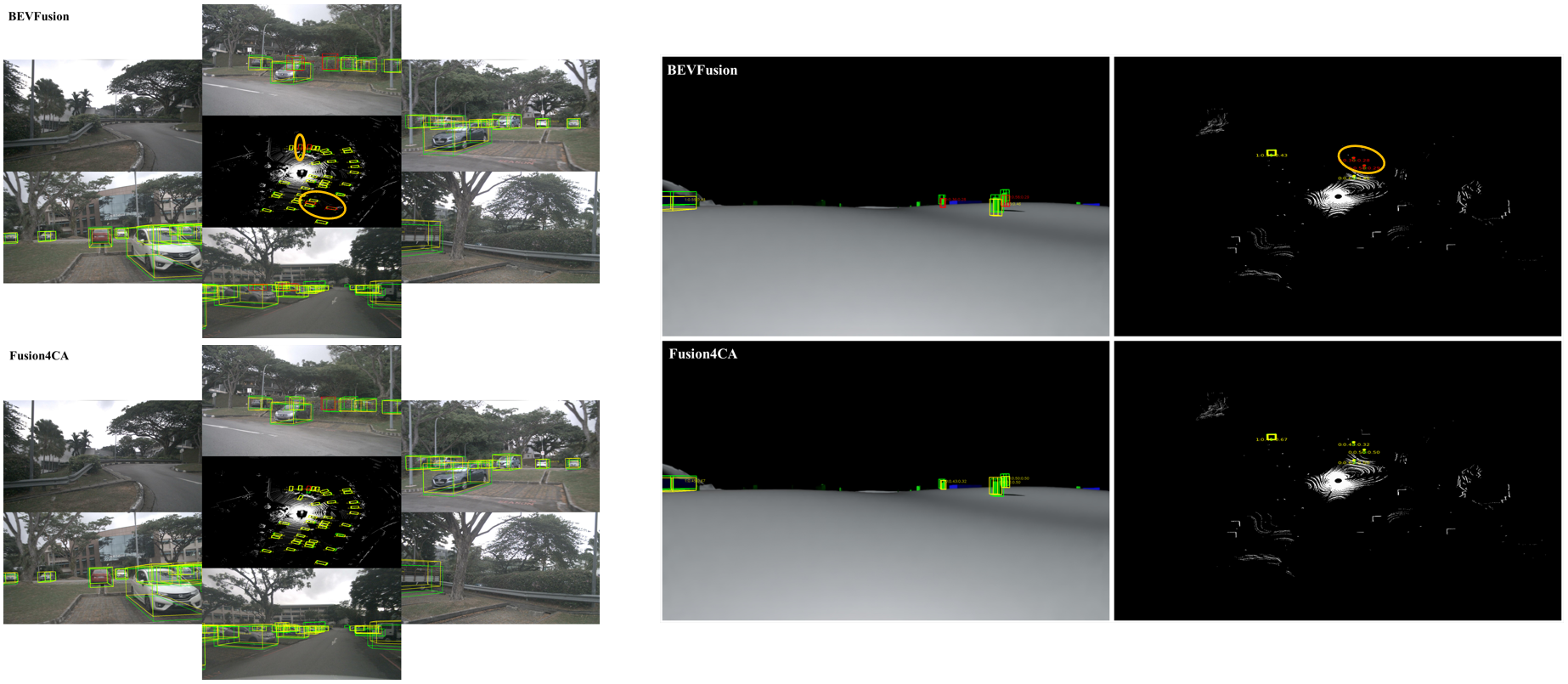}
        \caption{Visualization results between our method and the fully trained baseline.
        Green boxes denote ground truth, yellow boxes denote correct predictions, red boxes denote wrong predictions, and orange markers indicate instances correctly detected by our method but missed by the baseline.}
        \label{fig:Vis}
\end{figure*}

We evaluate our method on the nuScenes \cite{caesar2020nuscenes} validation and test sets for multi-class 3D object detection, with mAP and NDS as evaluation metrics. As shown in Table~\ref{tab:Comparision}, we compare our approach with several representative methods in recent years. Although our method is trained for only 6 epochs, which is considerably fewer than other competitors, it still outperforms them and achieves 69.7\% mAP and 72.1\% NDS. Moreover, compared with the fully trained multi-modal baseline \cite{liu2023bevfusion}, our method achieves 1.2\% mAP and 0.7\% NDS improvements on the validation set, and yields even greater performance gains over its LiDAR-only counterpart. Visualization can be seen in Fig.~\ref{fig:Vis} (left).
The results demonstrate the effectiveness of our method for 3D object detection in complex urban environments and validate that the proposed approach effectively exploits visual information from images.

\begin{table*}[htbp]
    \centering
    \caption{Comparision on simulated lunar dataset.
    }
    \label{tab:Comparasion_Sim}
    \begin{tabular}{c|c|cc|cc|ccc}
    \noalign{\hrule height 1pt}
    Method & Reference & mAP & NDS & Meteor & Platform & mATE & mASE & mAOE \\
    \hline 
    IS-Fusion \cite{yin2024fusion} & CVPR 2024 & 71.0 & 66.9 & 74.6 & 67.5 & 0.105 & 0.073 & 0.683 \\
    BEVFusion \cite{liu2023bevfusion} & ICRA 2023 & 88.8 & 81.6 & 84.9 & 92.8 & 0.096 & 0.043 & 0.146 \\
    \rowcolor{mybluebg}
    \textbf{Fusion4CA (Ours)} & - & \textbf{90.9} & \textbf{82.7} & \textbf{86.8} &\textbf{95.0} & \textbf{0.091} & \textbf{0.035} &\textbf{0.153} \\
    \noalign{\hrule height 1pt}
    \end{tabular}
\end{table*}

\subsection{Ablation Study on nuScenes Dataset}
To analyze the effects of different components, we train our model for 6 epochs on the nuScenes training set and conduct ablation experiments on the validation set, using mAP, NDS, and three average error metrics corresponding to translation, scale, and orientation. This study focuses on four key components: the Contrastive Alignment Module (ConAlign), the Camera Auxiliary Branch (CamAux), the Coordinate Attention Module (CoordAtt) and the Cognitive Adapter (CogAdp). 

As summarized in Table~\ref{tab:Ablation}, by comparing Order 01, 02, and 03, we observe that individually introducing either the Contrastive Alignment Module or the Camera Auxiliary Branch can substantially improve model performance. By comparing Order 06, 05, 04 and 01, we find that although individually adding the Coordinate Attention Module slightly degrades performance, combining it with other modules can further boost mAP from 68.9\% to 69.3\%. This phenomenon indirectly demonstrates that the auxiliary training modules help extract more effective information from the camera branch, which can then be further captured by the attention module. Furthermore, by incorporating the Cognitive Adapter and training with delta tuning, the proposed Fusion4CA (Order 07) achieves the best performance with 69.7\% mAP and 72.1\% NDS, improving by 5.0\% and 2.7\% respectively over the baseline (Order 01).

\subsection{Results in Simulated Lunar Environment}
Considering the relatively simple distribution of the simulated lunar environment, we train the model with 10 epochs to prevent potential overfitting and adopt a nuScenes-like evaluation protocol for consistent comparison. As reported in Table~\ref{tab:Comparasion_Sim}, our proposed method surpasses all competing approaches across various evaluation metrics, achieving 90.9\% mAP and 82.7\% NDS. Qualitative results are shown in Fig.~\ref{fig:Vis} (right). 
Notably, for the gray meteors (visualized in green), which share similar color and texture characteristics with the lunar surface, effective detection requires the camera modality to extract subtle visual cues and semantic features for accurate discrimination. Our method achieves 86.8\% mAP on this challenging category, surpassing the baseline by 1.9\%. This demonstrates the effectiveness of our approach in exploiting camera information, even under visually ambiguous conditions.
The superior performance under such limited training iterations and environment verifies the effective transferability and efficient exploitation of the camera modality, further confirming its practicality and adaptability in deployment scenarios.  

\section{CONCLUSION}
We propose Fusion4CA, a novel \textit{plug-and-play} Camera-LiDAR fusion framework that enhances BEV-based 3D object detection by fully exploiting RGB image information to address the over-reliance on LiDAR signals in existing multi-modal methods. Built upon BEVFusion, our framework integrates four complementary components to fully unleash the potential of visual inputs, including a Contrastive Alignment Module for geometric calibration of image features, a Camera Auxiliary Branch for supplementary supervision of the visual branch, a Cognitive Adapter \cite{yin20255} for efficient transfer of pre-trained image weights, and a Coordinate Attention module \cite{hou2021coordinate} for enhanced discriminative cross-modal fusion.
Remarkably, with only 6 training epochs, significantly fewer than conventional approaches, Fusion4CA outperforms the baseline by a notable margin while introducing only a minimal increase in inference parameters.
Extensive experiments conducted on nuScenes and simulated environment further demonstrate the effectiveness of our method.
This work provides a practical and efficient solution for autonomous driving, which fully exploits camera modality information and enables rapid transfer and deployment, thus advancing multi-modal 3D object detection in complex environments.


\bibliographystyle{IEEEtranBST/IEEEtran}
\bibliography{IEEEtranBST/ref}

@inproceedings{liu2023bevfusion,
  title={BEVFusion: Multi-Task Multi-Sensor Fusion with Unified Bird's-Eye View Representation},
  author={Liu, Zhijian and Tang, Haotian and Amini, Alexander and Yang, Xingyu and Mao, Huizi and Rus, Daniela and Han, Song},
  booktitle={IEEE International Conference on Robotics and Automation (ICRA)},
  year={2023}
}

@article{liang2022bevfusion,
  title={Bevfusion: A simple and robust lidar-camera fusion framework},
  author={Liang, Tingting and Xie, Hongwei and Yu, Kaicheng and Xia, Zhongyu and Lin, Zhiwei and Wang, Yongtao and Tang, Tao and Wang, Bing and Tang, Zhi},
  journal={Advances in neural information processing systems},
  volume={35},
  pages={10421--10434},
  year={2022}
}

@inproceedings{caesar2020nuscenes,
  title={nuscenes: A multimodal dataset for autonomous driving},
  author={Caesar, Holger and Bankiti, Varun and Lang, Alex H and Vora, Sourabh and Liong, Venice Erin and Xu, Qiang and Krishnan, Anush and Pan, Yu and Baldan, Giancarlo and Beijbom, Oscar},
  booktitle={Proceedings of the IEEE/CVF conference on computer vision and pattern recognition},
  pages={11621--11631},
  year={2020}
}

@inproceedings{philion2020lift,
  title={Lift, splat, shoot: Encoding images from arbitrary camera rigs by implicitly unprojecting to 3d},
  author={Philion, Jonah and Fidler, Sanja},
  booktitle={European conference on computer vision},
  pages={194--210},
  year={2020},
  organization={Springer}
}

@inproceedings{zhou2018voxelnet,
  title={Voxelnet: End-to-end learning for point cloud based 3d object detection},
  author={Zhou, Yin and Tuzel, Oncel},
  booktitle={Proceedings of the IEEE conference on computer vision and pattern recognition},
  pages={4490--4499},
  year={2018}
}

@article{yan2018second,
  title={Second: Sparsely embedded convolutional detection},
  author={Yan, Yan and Mao, Yuxing and Li, Bo},
  journal={Sensors},
  volume={18},
  number={10},
  pages={3337},
  year={2018},
  publisher={MDPI}
}

@article{zhao2024simplebev,
  title={Simplebev: Improved lidar-camera fusion architecture for 3d object detection},
  author={Zhao, Yun and Gong, Zhan and Zheng, Peiru and Zhu, Hong and Wu, Shaohua},
  journal={arXiv preprint arXiv:2411.05292},
  year={2024}
}

@inproceedings{bai2022transfusion,
  title={Transfusion: Robust lidar-camera fusion for 3d object detection with transformers},
  author={Bai, Xuyang and Hu, Zeyu and Zhu, Xinge and Huang, Qingqiu and Chen, Yilun and Fu, Hongbo and Tai, Chiew-Lan},
  booktitle={Proceedings of the IEEE/CVF conference on computer vision and pattern recognition},
  pages={1090--1099},
  year={2022}
}

@inproceedings{yin20255,
  title={5\%> 100\%: Breaking performance shackles of full fine-tuning on visual recognition tasks},
  author={Yin, Dongshuo and Hu, Leiyi and Li, Bin and Zhang, Youqun and Yang, Xue},
  booktitle={Proceedings of the Computer Vision and Pattern Recognition Conference},
  pages={20071--20081},
  year={2025}
}

@inproceedings{hou2021coordinate,
  title={Coordinate attention for efficient mobile network design},
  author={Hou, Qibin and Zhou, Daquan and Feng, Jiashi},
  booktitle={Proceedings of the IEEE/CVF conference on computer vision and pattern recognition},
  pages={13713--13722},
  year={2021}
}

@inproceedings{guan2022m3detr,
  title={M3detr: Multi-representation, multi-scale, mutual-relation 3d object detection with transformers},
  author={Guan, Tianrui and Wang, Jun and Lan, Shiyi and Chandra, Rohan and Wu, Zuxuan and Davis, Larry and Manocha, Dinesh},
  booktitle={Proceedings of the IEEE/CVF winter conference on applications of computer vision},
  pages={772--782},
  year={2022}
}

@article{arnold2019survey,
  title={A survey on 3d object detection methods for autonomous driving applications},
  author={Arnold, Eduardo and Al-Jarrah, Omar Y and Dianati, Mehrdad and Fallah, Saber and Oxtoby, David and Mouzakitis, Alex},
  journal={IEEE Transactions on Intelligent Transportation Systems},
  volume={20},
  number={10},
  pages={3782--3795},
  year={2019},
  publisher={IEEE}
}

@article{yurtsever2020survey,
  title={A survey of autonomous driving: Common practices and emerging technologies},
  author={Yurtsever, Ekim and Lambert, Jacob and Carballo, Alexander and Takeda, Kazuya},
  journal={IEEE access},
  volume={8},
  pages={58443--58469},
  year={2020},
  publisher={IEEE}
}

@article{li2020lidar,
  title={Lidar for autonomous driving: The principles, challenges, and trends for automotive lidar and perception systems},
  author={Li, You and Ibanez-Guzman, Javier},
  journal={IEEE Signal Processing Magazine},
  volume={37},
  number={4},
  pages={50--61},
  year={2020},
  publisher={IEEE}
}

@inproceedings{ibrahim2025multistream,
  title={Multistream Network for LiDAR and Camera-based 3D Object Detection in Outdoor Scenes},
  author={Ibrahim, Muhammad and Akhtar, Naveed and Wang, Haitian and Anwar, Saeed and Mian, Ajmal},
  booktitle={2025 IEEE/RSJ International Conference on Intelligent Robots and Systems (IROS)},
  pages={7796--7803},
  year={2025},
  organization={IEEE}
}

@inproceedings{wu2022sparse,
  title={Sparse fuse dense: Towards high quality 3d detection with depth completion},
  author={Wu, Xiaopei and Peng, Liang and Yang, Honghui and Xie, Liang and Huang, Chenxi and Deng, Chengqi and Liu, Haifeng and Cai, Deng},
  booktitle={Proceedings of the IEEE/CVF conference on computer vision and pattern recognition},
  pages={5418--5427},
  year={2022}
}

@inproceedings{wu2023virtual,
  title={Virtual sparse convolution for multimodal 3d object detection},
  author={Wu, Hai and Wen, Chenglu and Shi, Shaoshuai and Li, Xin and Wang, Cheng},
  booktitle={Proceedings of the IEEE/CVF conference on computer vision and pattern recognition},
  pages={21653--21662},
  year={2023}
}

@inproceedings{li2023bevdepth,
  title={Bevdepth: Acquisition of reliable depth for multi-view 3d object detection},
  author={Li, Yinhao and Ge, Zheng and Yu, Guanyi and Yang, Jinrong and Wang, Zengran and Shi, Yukang and Sun, Jianjian and Li, Zeming},
  booktitle={Proceedings of the AAAI conference on artificial intelligence},
  volume={37},
  number={2},
  pages={1477--1485},
  year={2023}
}

@inproceedings{lu2025toward,
  title={Toward real-world bev perception: Depth uncertainty estimation via gaussian splatting},
  author={Lu, Shu-Wei and Tsai, Yi-Hsuan and Chen, Yi-Ting},
  booktitle={Proceedings of the Computer Vision and Pattern Recognition Conference},
  pages={17124--17133},
  year={2025}
}

@article{li2024bevformer,
  title={Bevformer: learning bird’s-eye-view representation from lidar-camera via spatiotemporal transformers},
  author={Li, Zhiqi and Wang, Wenhai and Li, Hongyang and Xie, Enze and Sima, Chonghao and Lu, Tong and Yu, Qiao and Dai, Jifeng},
  journal={IEEE Transactions on Pattern Analysis and Machine Intelligence},
  volume={47},
  number={3},
  pages={2020--2036},
  year={2024},
  publisher={IEEE}
}

@inproceedings{yang2023bevformer,
  title={Bevformer v2: Adapting modern image backbones to bird's-eye-view recognition via perspective supervision},
  author={Yang, Chenyu and Chen, Yuntao and Tian, Hao and Tao, Chenxin and Zhu, Xizhou and Zhang, Zhaoxiang and Huang, Gao and Li, Hongyang and Qiao, Yu and Lu, Lewei and others},
  booktitle={Proceedings of the IEEE/CVF conference on computer vision and pattern recognition},
  pages={17830--17839},
  year={2023}
}

@inproceedings{jiang2023polarformer,
  title={Polarformer: Multi-camera 3d object detection with polar transformer},
  author={Jiang, Yanqin and Zhang, Li and Miao, Zhenwei and Zhu, Xiatian and Gao, Jin and Hu, Weiming and Jiang, Yu-Gang},
  booktitle={Proceedings of the AAAI conference on Artificial Intelligence},
  volume={37},
  number={1},
  pages={1042--1050},
  year={2023}
}

@inproceedings{qi2017pointnet,
  title={Pointnet: Deep learning on point sets for 3d classification and segmentation},
  author={Qi, Charles R and Su, Hao and Mo, Kaichun and Guibas, Leonidas J},
  booktitle={Proceedings of the IEEE conference on computer vision and pattern recognition},
  pages={652--660},
  year={2017}
}

@article{qi2017pointnet++,
  title={Pointnet++: Deep hierarchical feature learning on point sets in a metric space},
  author={Qi, Charles Ruizhongtai and Yi, Li and Su, Hao and Guibas, Leonidas J},
  journal={Advances in neural information processing systems},
  volume={30},
  year={2017}
}

@inproceedings{ding2019votenet,
  title={Votenet: A deep learning label fusion method for multi-atlas segmentation},
  author={Ding, Zhipeng and Han, Xu and Niethammer, Marc},
  booktitle={International conference on medical image computing and computer-assisted intervention},
  pages={202--210},
  year={2019},
  organization={Springer}
}

@inproceedings{lang2019pointpillars,
  title={Pointpillars: Fast encoders for object detection from point clouds},
  author={Lang, Alex H and Vora, Sourabh and Caesar, Holger and Zhou, Lubing and Yang, Jiong and Beijbom, Oscar},
  booktitle={Proceedings of the IEEE/CVF conference on computer vision and pattern recognition},
  pages={12697--12705},
  year={2019}
}

@article{hu2023ea,
  title={Ea-lss: Edge-aware lift-splat-shot framework for 3d bev object detection},
  author={Hu, Haotian and Wang, Fanyi and Su, Jingwen and Wang, Yaonong and Hu, Laifeng and Fang, Weiye and Xu, Jingwei and Zhang, Zhiwang},
  journal={arXiv preprint arXiv:2303.17895},
  year={2023}
}

@inproceedings{qi2018frustum,
  title={Frustum pointnets for 3d object detection from rgb-d data},
  author={Qi, Charles R and Liu, Wei and Wu, Chenxia and Su, Hao and Guibas, Leonidas J},
  booktitle={Proceedings of the IEEE conference on computer vision and pattern recognition},
  pages={918--927},
  year={2018}
}

@inproceedings{vora2020pointpainting,
  title={Pointpainting: Sequential fusion for 3d object detection},
  author={Vora, Sourabh and Lang, Alex H and Helou, Bassam and Beijbom, Oscar},
  booktitle={Proceedings of the IEEE/CVF conference on computer vision and pattern recognition},
  pages={4604--4612},
  year={2020}
}

@article{cai2023bevfusion4d,
  title={Bevfusion4d: Learning lidar-camera fusion under bird's-eye-view via cross-modality guidance and temporal aggregation},
  author={Cai, Hongxiang and Zhang, Zeyuan and Zhou, Zhenyu and Li, Ziyin and Ding, Wenbo and Zhao, Jiuhua},
  journal={arXiv preprint arXiv:2303.17099},
  year={2023}
}

@inproceedings{zhang2024sparselif,
  title={SparseLIF: High-performance sparse LiDAR-camera fusion for 3D object detection},
  author={Zhang, Hongcheng and Liang, Liu and Zeng, Pengxin and Song, Xiao and Wang, Zhe},
  booktitle={European conference on computer vision},
  pages={109--128},
  year={2024},
  organization={Springer}
}

@article{wang2025mv2dfusion,
  title={Mv2dfusion: Leveraging modality-specific object semantics for multi-modal 3d detection},
  author={Wang, Zitian and Huang, Zehao and Gao, Yulu and Wang, Naiyan and Liu, Si},
  journal={IEEE Transactions on Pattern Analysis and Machine Intelligence},
  year={2025},
  publisher={IEEE}
}

@inproceedings{chen2020simple,
  title={A simple framework for contrastive learning of visual representations},
  author={Chen, Ting and Kornblith, Simon and Norouzi, Mohammad and Hinton, Geoffrey},
  booktitle={International conference on machine learning},
  pages={1597--1607},
  year={2020},
  organization={PmLR}
}

@inproceedings{yin2021center,
  title={Center-based 3d object detection and tracking},
  author={Yin, Tianwei and Zhou, Xingyi and Krahenbuhl, Philipp},
  booktitle={Proceedings of the IEEE/CVF conference on computer vision and pattern recognition},
  pages={11784--11793},
  year={2021}
}

@article{yadav2019evalai,
  title={Evalai: Towards better evaluation systems for ai agents},
  author={Yadav, Deshraj and Jain, Rishabh and Agrawal, Harsh and Chattopadhyay, Prithvijit and Singh, Taranjeet and Jain, Akash and Singh, Shiv Baran and Lee, Stefan and Batra, Dhruv},
  journal={arXiv preprint arXiv:1902.03570},
  year={2019}
}

@inproceedings{chen2022focal,
  title={Focal sparse convolutional networks for 3d object detection},
  author={Chen, Yukang and Li, Yanwei and Zhang, Xiangyu and Sun, Jian and Jia, Jiaya},
  booktitle={Proceedings of the IEEE/CVF conference on computer vision and pattern recognition},
  pages={5428--5437},
  year={2022}
}

@inproceedings{wang2021pointaugmenting,
  title={Pointaugmenting: Cross-modal augmentation for 3d object detection},
  author={Wang, Chunwei and Ma, Chao and Zhu, Ming and Yang, Xiaokang},
  booktitle={Proceedings of the IEEE/CVF conference on computer vision and pattern recognition},
  pages={11794--11803},
  year={2021}
}

@inproceedings{song2023graphalign,
  title={GraphAlign: Enhancing accurate feature alignment by graph matching for multi-modal 3D object detection},
  author={Song, Ziying and Wei, Haiyue and Bai, Lin and Yang, Lei and Jia, Caiyan},
  booktitle={Proceedings of the IEEE/CVF international conference on computer vision},
  pages={3358--3369},
  year={2023}
}

@article{yin2021multimodal,
  title={Multimodal virtual point 3d detection},
  author={Yin, Tianwei and Zhou, Xingyi and Kr{\"a}henb{\"u}hl, Philipp},
  journal={Advances in Neural Information Processing Systems},
  volume={34},
  pages={16494--16507},
  year={2021}
}

@inproceedings{zhang2024safdnet,
  title={Safdnet: A simple and effective network for fully sparse 3d object detection},
  author={Zhang, Gang and Chen, Junnan and Gao, Guohuan and Li, Jianmin and Liu, Si and Hu, Xiaolin},
  booktitle={Proceedings of the IEEE/CVF conference on computer vision and pattern recognition},
  pages={14477--14486},
  year={2024}
}

@inproceedings{chen2023voxelnext,
  title={Voxelnext: Fully sparse voxelnet for 3d object detection and tracking},
  author={Chen, Yukang and Liu, Jianhui and Zhang, Xiangyu and Qi, Xiaojuan and Jia, Jiaya},
  booktitle={Proceedings of the IEEE/CVF conference on computer vision and pattern recognition},
  pages={21674--21683},
  year={2023}
}

@inproceedings{xu2021fusionpainting,
  title={Fusionpainting: Multimodal fusion with adaptive attention for 3d object detection},
  author={Xu, Shaoqing and Zhou, Dingfu and Fang, Jin and Yin, Junbo and Bin, Zhou and Zhang, Liangjun},
  booktitle={2021 IEEE international intelligent transportation systems conference (ITSC)},
  pages={3047--3054},
  year={2021},
  organization={IEEE}
}

@inproceedings{chen2023futr3d,
  title={Futr3d: A unified sensor fusion framework for 3d detection},
  author={Chen, Xuanyao and Zhang, Tianyuan and Wang, Yue and Wang, Yilun and Zhao, Hang},
  booktitle={proceedings of the IEEE/CVF conference on computer vision and pattern recognition},
  pages={172--181},
  year={2023}
}

@inproceedings{yin2024fusion,
  title={Is-fusion: Instance-scene collaborative fusion for multimodal 3d object detection},
  author={Yin, Junbo and Shen, Jianbing and Chen, Runnan and Li, Wei and Yang, Ruigang and Frossard, Pascal and Wang, Wenguan},
  booktitle={Proceedings of the IEEE/CVF conference on computer vision and pattern recognition},
  pages={14905--14915},
  year={2024}
}

\end{document}